\documentclass{article}
\usepackage{graphicx}
\graphicspath{ {./} }
\usepackage[utf8]{inputenc}
\usepackage{listings}
\usepackage{xcolor}
\usepackage{caption}
\usepackage{subcaption}
\usepackage{graphicx}

\title{Recognizing Exercises and Counting Repetitions in Real Time}
\author{Talal Alatiah, Chen Chen \\
Department of Electrical and Computer Engineering \\
University of North Carolina at Charlotte}

\date{May 2020}

\begin{document}

\maketitle
\begin{abstract}
Artificial intelligence technology has made its way absolutely necessary in a variety of industries including the fitness industry. Human pose estimation is one of the important researches in the field of Computer Vision for the last few years. In this project, pose estimation and deep machine learning techniques are combined to analyze the performance and report a feedback on the repetitions of performed exercises in real time. Involving machine learning technology in fitness industry could help the judges to count repetitions of any exercise during Weightlifting or CrossFit competitions.
\end{abstract}

\section{Introduction}
The project provides a solution to count repetitions of a physical exercise in real time. The method uses pose estimation to track athletes, recognize their performed exercises, count the repetitions, and analyze the performance of the repetitions. OpenPose~\cite{openpose}  is a real-time network that detects human poses and extracts their 3D skeleton keypoints from an input video or an external camera. The proposed method uses a pretrained model on BODY\_25 dataset for OpenPose, which is faster and more accurate comparing with MSCOCO dataset~\cite{openposeFAQ}. The method uses these keypoints to track poses on frames and recognize the performed exercise. A Set of exercise videos are used from UCF101 dataset ~\cite{UCF101} to train the exercise recognition model. In order to an effective counting and analyzing repetitions, each exercise has different pre-selected parameters; upper and lower range of motion, major joint, and type of motion. The method measures, filters, and smooths the angles of the major joint for the performed exercise. Then, it counts the repetitions of the exercise, and detects the number of correct and incorrect repetitions, and locates their frames.\\

In Crossfit competitions, each athlete is paired with an individual judge. The judge counts and evaluates the athlete’s repetitions. In large competitions, judges are limited to count repetitions without evaluation~\cite{judge}. Involving deep learning technology in this field could eliminate the need of this number of judges. This might reduce the budget of judges’ salaries during these competitions, as well as solving the limitation of judges.\\

\begin{figure}
\includegraphics{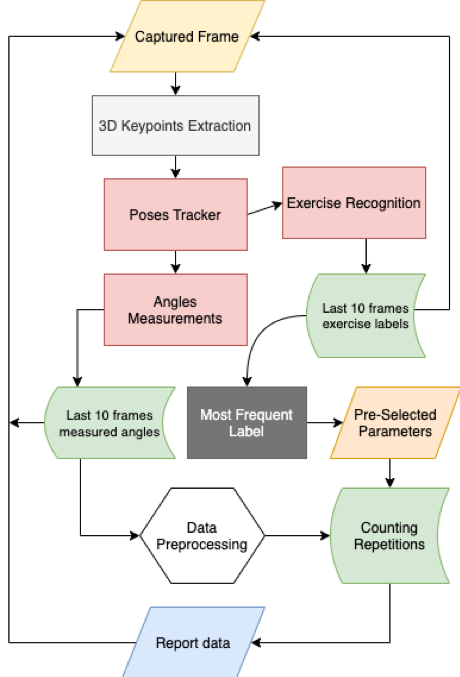}
\centering

\caption{The proposed method diagram}
\label{fig:diagram}
\end{figure}

The challenge of existing publicly available datasets of video recordings with its repetitions for CrossFit was concerned in this project. So, a total of 25 videos were recorded and others were downloaded from YouTube. Then, the performances of the exercise were evaluated manually to test the proposed method.\\

The proposed method achieved 98.4\% in the accuracy of exercise recognition, which is high ratio and closes to highest ratio of the related works. In addition, it achieved an error of counter within \( \pm \) 1 reps, which similar to the best result of related works. However, the proposed method is the only method that can recognize and report the incorrect from the total repetitions. In addition, the model counts the total repetitions for each exercise or sets separately if a person performs multiple exercises and keeps shifting between them.\\

The main contributions of this project are:
\begin{itemize}
  \item Offering a complete system to re-identify and track persons, recognize the performed exercises, and count each person’s repetitions in real time.
  \item Analyzing the performed exercises without the requirement of recording the video from a specific particular perspective.
  \item Distinguishing and reporting total, correct, and incorrect repetitions.
\end{itemize}

\section{Related Works}
There are some existed methods that offer solutions to recognize exercise and/or counting repetitions: using either a smartwatch, a security camera system, or a webcam.

\subsection{Smartwatch}
In paper~\cite{smartwatch}, a deep learning approach is constructed for exercise recognition and repetition counting. The method uses a raw sensor data extracted from a smartwatch to train a neural network. It achieves a classification accuracy of 99.96\% and counting correctly within an error of \( \pm \) 1 repetition in 91\% of the performed sets. In order to apply the method, every athlete must have a smartwatch, which requires a large budget. In addition, the method is not able to indicate correct/incorrect repetitions.

\subsection{GymCam}
GymCam~\cite{gymcam} is a camera-based system for detecting, recognizing and tracking preformed exercises. The system uses recorded gym video to train a neural network. The method recognizes exercises with an accuracy of 93.6\% and counting the number of repetitions within \( \pm \) 1.7 on average. It’s able to recognize and count the repetition of multiple persons appearing in the frame. Similar to the previous method [9], GymCam is not able to indicate correct/incorrect repetitions.

\subsection{Pose Trainer}
Pose Trainer~\cite{posetrainer} uses pose estimation method to detect the athlete’s exercise pose and provides in details a useful feedback. The model relies on recording a dataset of over 100 exercise videos of correct and incorrect form, based on personal training guidelines, and build geometric-heuristic and machine learning method for evaluation. This method also is not able to analyze the performance of multiple persons appearing in the frame, as well as it not able to count the repetition of the performed exercise. It’s also limited to four exercises; bicep curl, front raises, shrugs, and shoulder press. Another disadvantage is that the user is required to record the video from a particular perspective (facing the camera, side to the camera, etc.)

\section{Proposed Method}
The proposed method divides respectively into three phases; pose tracker to identify and track athlete to apply the algorithm to multiple persons, exercise recognition to detect the name of the appeared exercises, and counter to count and indicate the correct and incorrect repetitions. Figure \ref{fig:diagram} sketches a flowchart of the proposed method.

\subsection{Pose Tracker}
\begin{figure}[ht]
\includegraphics{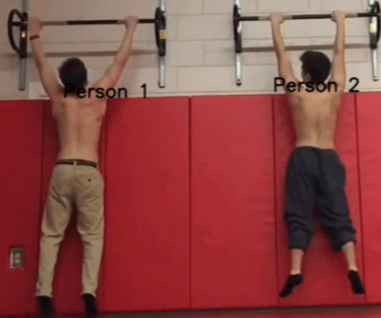}
\centering

\caption{Appling Person Re-Identification technique with labeling}
\label{fig:re-identification}
\end{figure}

The proposed method identifies and tracks each person appeared in the frames to analyze/count separately his repetitions. For this task, the model uses Euclidean distance score for each 3D body key-point appeared in the current and previous frames. Then, the system rearranges and assigns each person to his unique ID number. Figure \ref{fig:re-identification} shows an example of implementation of person re-identification technique for two persons appearing the videos.

\begin{figure}[ht]
\includegraphics[width=\textwidth]{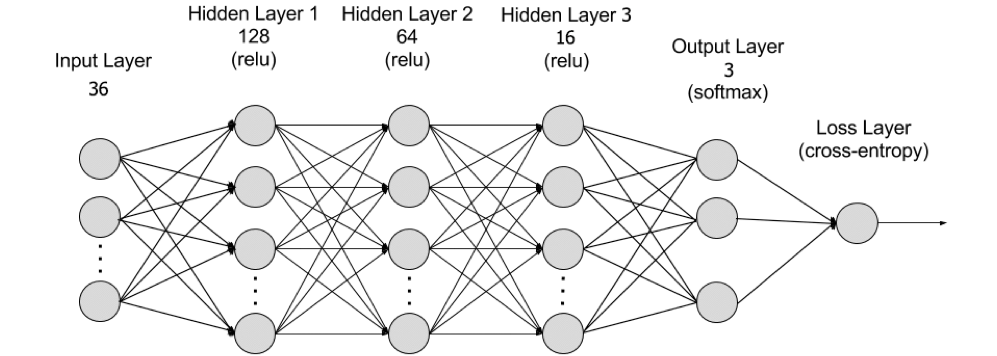}
\centering

\caption{Exercise recognition neural network architecture} 
\label{fig:exercise_recognition_neural_network}
\end{figure}

\subsection{Exercise Recognition}
\begin{figure}[ht]
\includegraphics{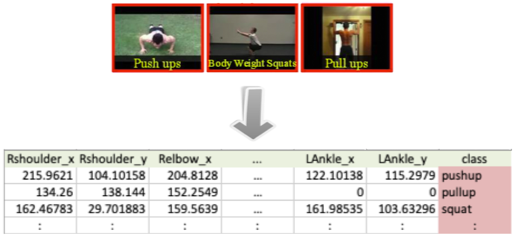}
\centering

\caption{Data preprocessing for exercise recognition}
\label{fig:data_preprocessing}
\end{figure}

Basically, it’s an action recognition model that uses 103 push-ups, 101 pull-ups, and 113 squats videos from UCF101 dataset~\cite{UCF101}. The model uses OpenPose~\cite{openpose} to extract 3D human body keypoints from each frame. Then, the keypoints are gotten rid of undetected data, and convert from 3D to 2D mesh as shown in figure \ref{fig:data_preprocessing}. These keypoints were considered as features to classify the performed exercises. The final dataset that uses to train the action recognition neural network contains 7437 push-ups, 12707 pull-ups, and 14357 squats body keypoints. Figure \ref{fig:exercise_recognition_neural_network} demonstrates the architecture of the exercise recognition neural network. In order to detect the known and unknown pose/motion,  the model classifies pose with reject option~\cite{rejectoption} using confidence interval thresholds. The mean of the Softmax probabilities from the test set of each class was collected and the 90\% confidence interval was estimated to be a reference on final model. The predicted probabilities for each class of the exercise recognition network must be bounded between these estimated confidence intervals to confirm the predicted label or will label as an unknown. 

\subsection{Repetitions Counter}
For this task, OpenPose extracts 3D body keypoints from an input videos or an external webcam. The method recognizes exercise in each frame. After passing 10 frames, the system takes the most frequent prediction from last 10 frames to figure out the performed exercise. Eq.\ref{performedexercise} shows the formula for this process.
\begin{equation}
performed\ exercise(f) = most\_common(y(f-9),y(f-8),...,y(f))
\label{performedexercise}
\end{equation}
Where, f: frame, and y: predicted label from the network.\\

After indicating the label of the performed exercise in the frame, the model will use preselected parameters for each exercise. These parameters contain the exercise range of motion, the major joint, and type of motion (push or pull). Then, the system calculates the angles of the major joint using vector dot product~\cite{lawofcosine} as shown in the following equations.

\begin{equation}
\overrightarrow{BA} = A-B
\end{equation}
\begin{equation}
\overrightarrow{BC} = C-B
\end{equation}
\begin{equation}
\Theta =\arccos(\frac{\overrightarrow{BA}\cdot \overrightarrow{BC}}{\left \|\overrightarrow{BC}\right \| \left \|\overrightarrow{BA}\right \|})
\end{equation}
Where, A, B, and C: 3D points.
\subsubsection{Data Preprocessing}
After measuring each joint angle, the system applies data preprocessing techniques to the calculated angles: filling the gaps and normalizing the outliers.\\

To fill the gaps, increment/decrement operators are applied as shown in Eq.\ref{gaps}.
\begin{equation}
\angle _{f}=\left |\angle _{f-1} -(\angle _{f-2} - \angle _{f-1} ))\right | = \left |2\ \angle _{f-2} - \angle _{f-1} \right |
\label{gaps}
\end{equation}
Where, \angle: The angles, and f: Frame.\\

To normalizing the outliers, the following proposed formula is applied for N iterations.
\begin{equation}
\mbox{\fontsize{9.5}{9.5}\selectfont\( %
\angle _{f} = \frac{\angle _{f-1}+\angle _{f+1}}{2}\ \ if(\angle _{f} > M\ \& \ \angle _{f} < \frac{\angle _{f-1}+\angle _{f+1}}{2}) or(\angle _{f} < M\ \& \ \angle _{f} > \frac{\angle _{f-1}+\angle _{f+1}}{2})
\)} %
\end{equation}
Where, \angle: The angles, f: Frame, and M: Middle of range of motion.

\subsubsection{Counting Algorithm}
\begin{figure}[ht]
\includegraphics[width=0.5\textwidth]{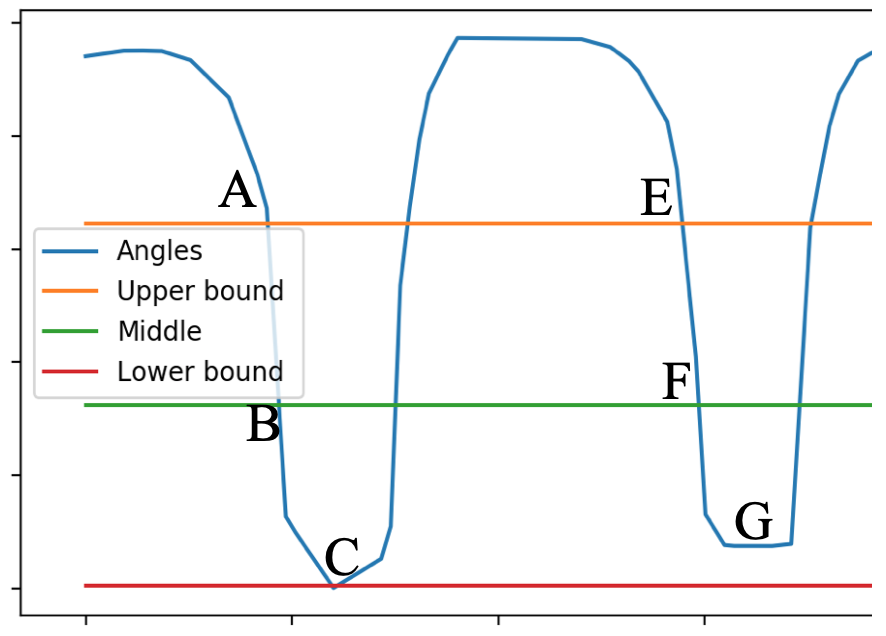}
\centering

\caption{Sample of two cycles (repetitions).}
\label{fig:cyclerep}
\end{figure}

Figure \ref{fig:cyclerep} represents two cycles (repetitions): A-C (1st cycle) and E-G (2nd cycle). Simply, if the angle passes the middle line of the range of motion (point B), this will count as one repetition. If the following angles (trend) pass the upper/lower of the range of motion (point C) and continue pass the opposite side of the range of motion (point E), this will count as a correct repetition. Otherwise, it will count as an incorrect repetition (point G). So, the first cycle will be count as correct repetitions, while the second one is incorrect.

\subsection{Real-time Implementation}
One of the major considerations in this project is achievement of run the model in real-time. The essential challenge was achieving higher frame per second (FPS) value using an available computer hardware and equipment. The proposed model is run and tested in a MacBook Pro NVIDIA GeForce GT 750M 2 GB, and an integrated 720p FaceTime HD webcam. The hardware is limited and inefficient to implement deep learning models. However, the keypoints extraction using OpenPose model is executed in Google Colab~\cite{googlecolab}, a free cloud service that supports free GPU. Hence, the proposed method tested using a prerecorded video. Table \ref{tab:cpu} demonstrates results of two attempts to implement real-time OpenPose model using different dataset on CPU of the MacBook Pro.

\begin{table}[h]
\centering
\begin{tabular}{|r|c|c|}
\hline
\multicolumn{1}{|c|}{\textbf{Dataset}} & \textbf{FPS}       & \textbf{Accuracy} \\ \hline
BODY\_25~\cite{openpose}             & $\sim$ 0.6 & High     \\ \hline
MobileNet-thin~\cite{poseestimationtf}          & $\sim$ 4.2 & Low      \\ \hline
\end{tabular}
\caption{Two attempts to implement real-time pose estimation using CPU.}
\label{tab:cpu}
\end{table}

MobileNet-thin model only extracts 2D keypoints. So, there was an attempt to use 3D pose baseline model as a medium to convert keypoints from 2D to 3D~\cite{3dbaseline}. however, the model used a lot of RAM storage, which made this attempt inefficient.\\

Due to the achievement of these low values of FPS using CPU, the proposed method was built and evaluated using prerecorded videos within fps of 24 - 30. However, the final system was ran and tested on GPU using jetson AGX Xavier which achieved a FPS of $\sim$ 30.

\section{Results}
Since the model, divided into three main tasks, each task has its own result. However, the pose tracker was not evaluated in the current model.

\subsection{Exercise Recognition}
\begin{figure}[ht]
\includegraphics[width=0.5\textwidth]{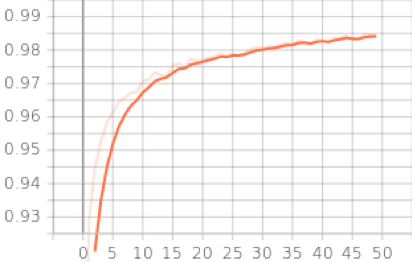}
\centering

\caption{Plot of the model accuracy on training set.}
\label{fig:accuracy}
\end{figure}

\begin{figure}[ht]
\includegraphics[width=0.5\textwidth]{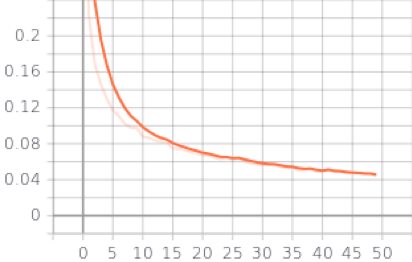}
\centering
\caption{Plot of the model loss on training set.}
\label{fig:loss}
\end{figure}

Figures \ref{fig:accuracy} and \ref{fig:loss} visualize the performance of the exercise recognition neural network during training process over 50 epochs. Tables \ref{tab:accuracy} and \ref{tab:class-wise} represent the accuracy of the proposed method compared with the related works and class-wise accuracy. Since the related works use different methods and datasets, comparing models in a fair manner is often not exist. However, exercise recognition is a partial task of the final project.

\begin{table}[!ht]
\centering
\begin{tabular}{|l|c|}
\hline
\textbf{Method}          & \textbf{Accuracy}         \\ \hline
Smartwatch      & \textbf{99.96\%} \\ \hline
GymCam          & 93.6\%           \\ \hline
Proposed Method & 98.4\%           \\ \hline       
\end{tabular}
\caption{The accuracy of exercise recognition.}
\label{tab:accuracy}
\end{table}
\begin{table}[!ht]
\centering
\begin{tabular}{|r|c|c|c|}
\hline
\multicolumn{1}{|c|}{\textbf{Class}} & \textbf{Precision} & \textbf{Recall} & \textbf{F1-Score} \\ \hline
Pull up                              & 0.975              & 0.982           & 0.979             \\ \hline
Push up                              & 0.975              & 0.968           & 0.972             \\ \hline
Squat                                & \textbf{0.992}     & \textbf{0.989}  & \textbf{0.990}    \\ \hline
\end{tabular}
\caption{Class-wise accuracy.}
\label{tab:class-wise}
\end{table}

\subsection{Repetitions Counter}
The following sections represent samples of preprocessed data (angles), and the accuracy of counting repetitions for each method.

\subsubsection{Data Preprocessing}
\begin{figure}[ht]
\includegraphics[width=\textwidth]{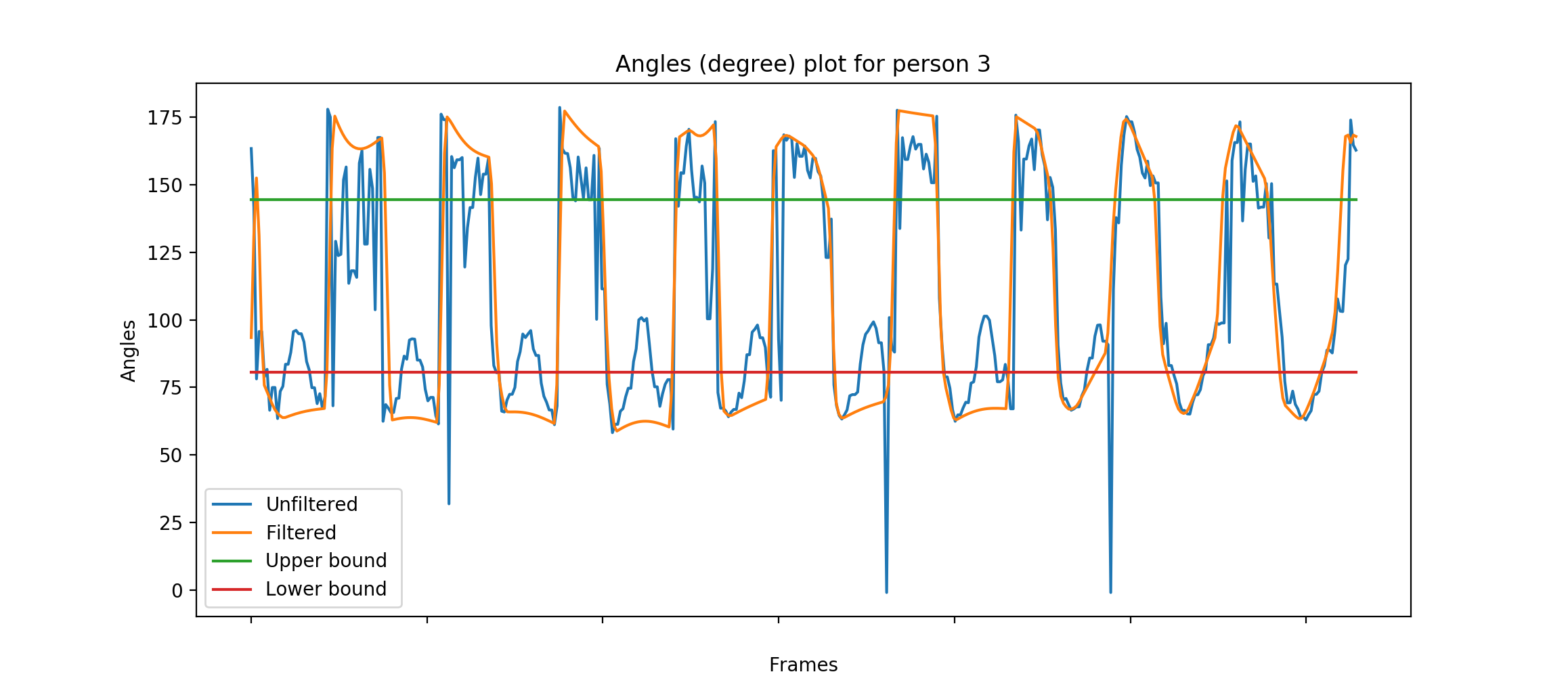}
\includegraphics[width=\textwidth]{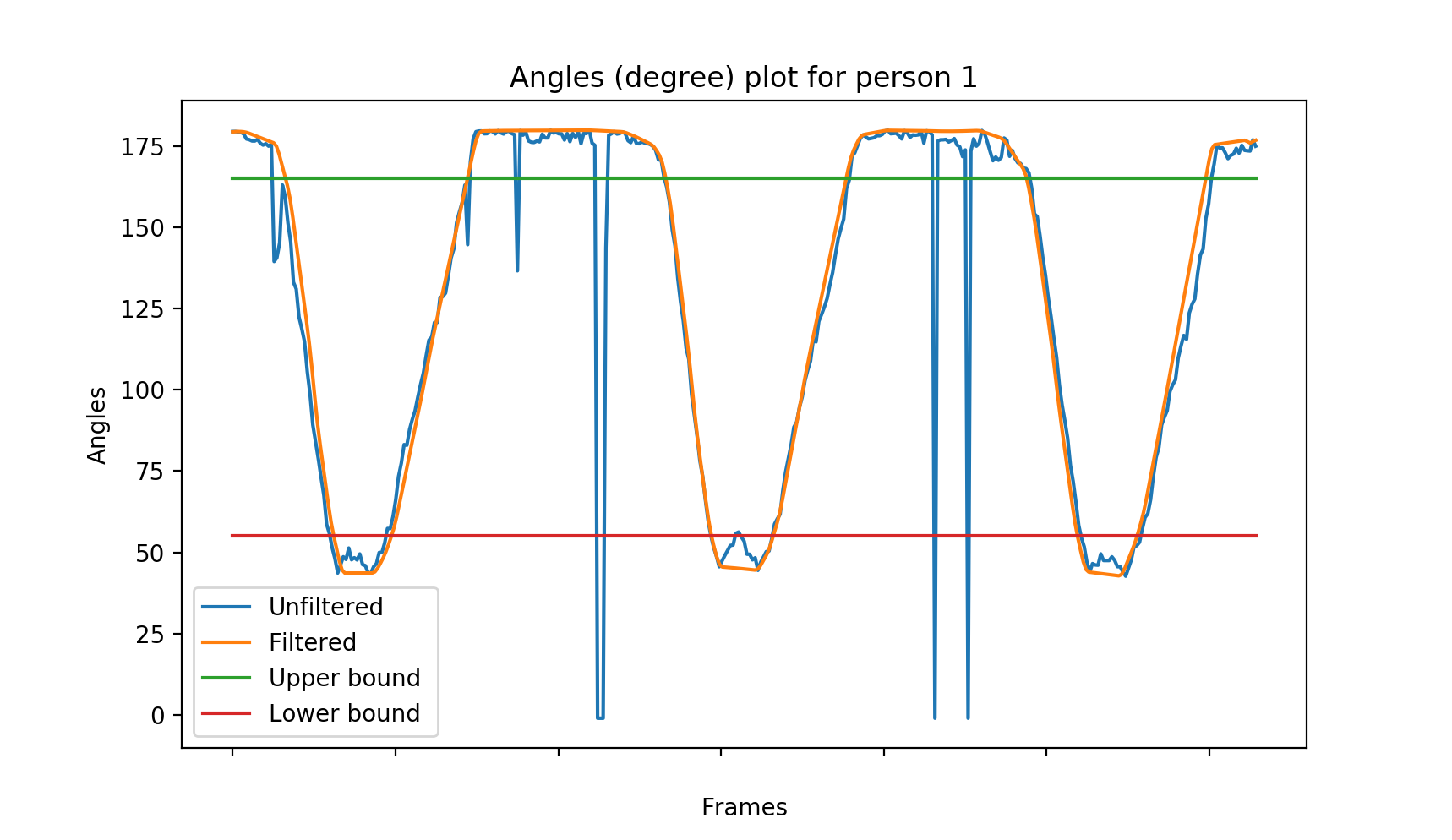}
\centering

\caption{Two examples of data preprocessing}
\label{fig:data_preprocessing1}
\end{figure}

Figure \ref{fig:data_preprocessing1} demonstrates two graphs of unfiltered and filtered angles. The model filled the gaps and normalized the outliers from the signal (angles).

\subsubsection{Counting Repetitions}
Table \ref{tab:counter} represents the counter error of the proposed method compared with the related works. This comparison as well is not fair due to limited dataset. In addition, there is no existed method that counts correct and incorrect repetition. It is notable that correct/incorrect counting in the proposed method depends on the preselected parameters of the exercise.

\begin{table}[!h]
\centering
\begin{tabular}{|c|c|}
\hline
\textbf{Method} & \textbf{Error} \\ \hline
Smartwatch      & \( \pm \) 1 rep        \\ \hline
GymCam          & \( \pm \) 1.7   reps   \\ \hline
Proposed Method & \( \pm \) 1 rep        \\ \hline
\end{tabular}
\caption{The error of counter}
\label{tab:counter}
\end{table}

\subsection{Output Format}
\begin{figure}[ht]
\includegraphics[width=0.30\textwidth]{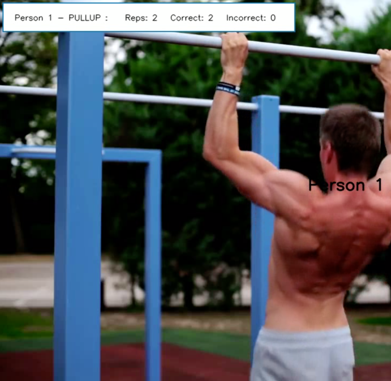}
\centering

\caption{Snapshot of an output of a labeled video.}
\label{fig:labeledvideo}
\end{figure}

The system generates two formats of an output. First output is a video with label of total, correct, and incorrect repetitions as shown in figure \ref{fig:labeledvideo}. The second output is a text report of each indicated repetition with its time in the original video. Below is an example of text report output on a correctly and incorrectly performed push-up exercise:

\begin{lstlisting}
One Attempt! - 00:19 sec
One Attempt! - 01:29 sec
One Attempt! - 02:27 sec
- Correct! - 03:51 sec
One Attempt! - 04:34 sec
One Attempt! - 06:41 sec
- Correct! - 09:16 sec
One Attempt! - 09:43 sec
- Correct! - 10:58 sec
One Attempt! - 11:20 sec

Person 1
Predicted Exercise: push-up
Total Reps:  7
Correct Reps:  3
Incorrect Reps:  4

\end{lstlisting}

\section{Conclusion and Future Work}
Involving deep learning technology in this field could help the judges during sport competitions. However, currently, I believe that the human decision is still do better than computer. For the future work, the system will be developed to analyze complex exercises, which use multiple joints to perform. The current method is unable to distinguish between the competitors and judges. It’s just ignored any known movement for any person appeared in the frames. In future work, the system will be developed to identify and ignore the judges and audience. Another work is to provide a contribution of a public dataset of videos with its repetitions for CrossFit.

\begin{figure}[ht]
\centering
\begin{subfigure}{.5\textwidth}
  \centering
  \includegraphics[width=0.85\linewidth]{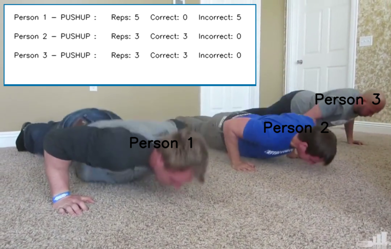}
  \caption{Push up contest (3 persons)}
  \label{fig:sub1}
\end{subfigure}%
\begin{subfigure}{.5\textwidth}
  \centering
  \includegraphics[width=0.55\linewidth]{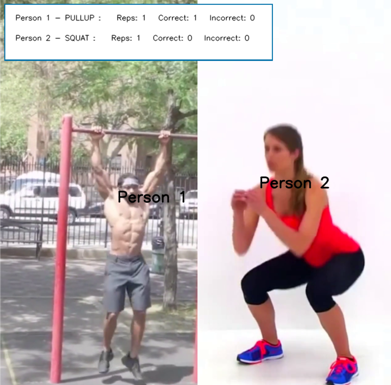}
  \caption{Two different actions (2 persons)}
  \label{fig:sub2}
\end{subfigure}
\caption{Snapshots of multiple persons in video.}
\end{figure}

\begin{figure}[ht]
\centering
\begin{subfigure}{.5\textwidth}
  \centering
  \includegraphics[width=0.85\linewidth]{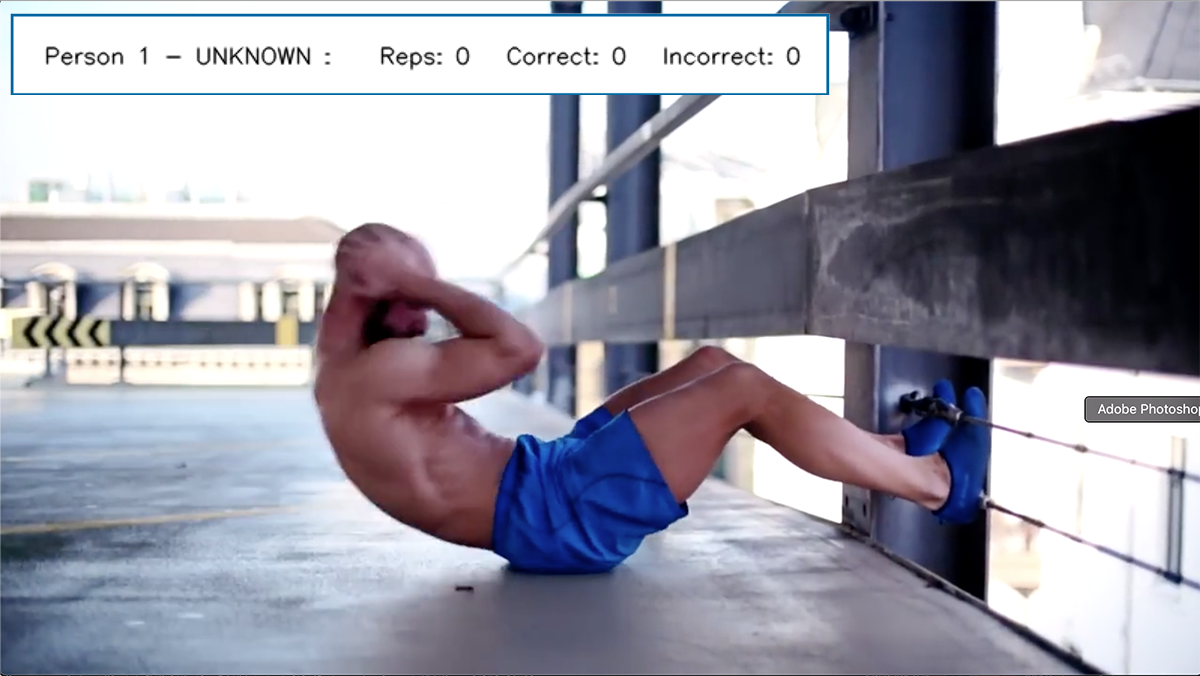}
  \caption{Sit-up (Unknown)}
\end{subfigure}%
\begin{subfigure}{.5\textwidth}
  \centering
  \includegraphics[width=0.85\linewidth]{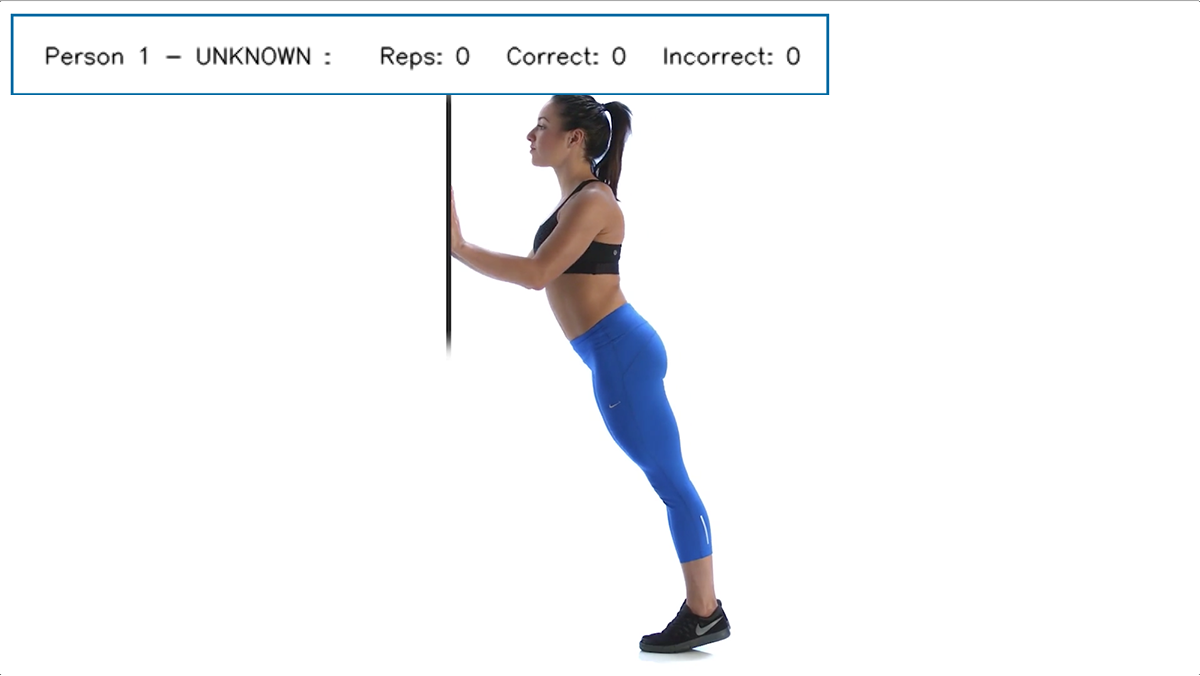}
  \caption{Wall push-up (Unknown)}
\end{subfigure}
\caption{Snapshots of unknown action or exercise.}
\label{fig:unknown_action}
\end{figure}

\begin{figure}
\includegraphics[width=\textwidth]{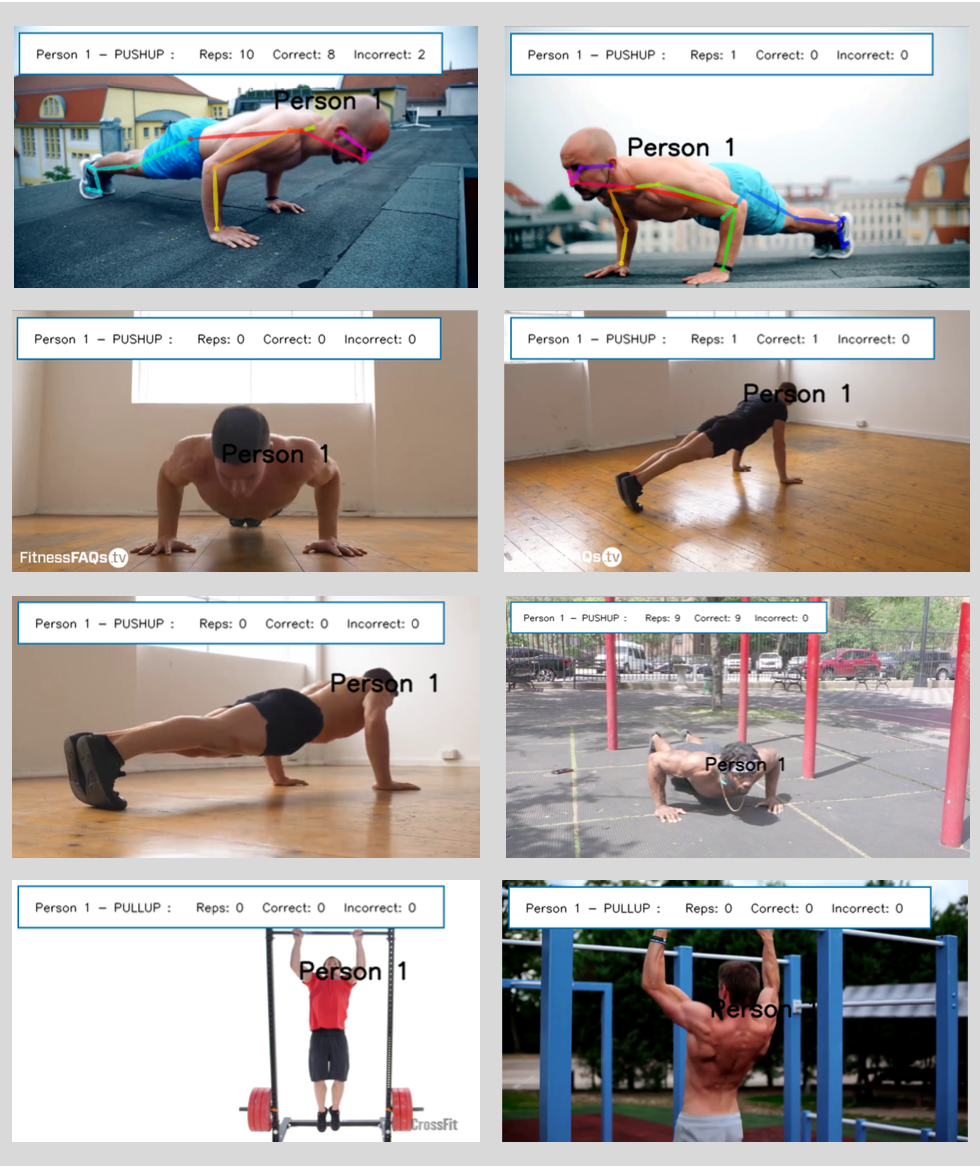}
\centering

\caption{Outputs of different camera angles.}
\label{fig:different_camera_angles}
\end{figure}

\pagebreak

\end{document}